\documentclass{article}

\PassOptionsToPackage{numbers,compress,sort}{natbib}

\usepackage[preprint,eandd]{neurips_2026}

\usepackage[utf8]{inputenc}
\usepackage[T1]{fontenc}
\usepackage{hyperref}
\usepackage{url}
\usepackage{graphicx}
\usepackage{booktabs}
\usepackage{multirow}
\usepackage{amsfonts}
\usepackage{amsmath}
\usepackage{amssymb}
\usepackage{nicefrac}
\usepackage{microtype}
\usepackage{xcolor}
\usepackage{colortbl}
\usepackage{algorithm}
\usepackage{algpseudocode}
\usepackage{float}
\usepackage{listings}
\usepackage{enumitem}
\usepackage{tikz}
\usetikzlibrary{positioning, arrows.meta, fit, calc, shapes.geometric, patterns}
\usepackage{pgfplots}
\pgfplotsset{compat=1.18}

\title{BioMedArena: An Open-source Toolkit for Building and Evaluating Biomedical Deep Research Agents}

\author{
    Jinge Wu\thanks{Equal contribution.}, 
  \ Hongjian Zhou\footnotemark[1], 
  \ Mingde Zeng,
  \ Jiayuan Zhu, 
  \ Junde Wu, \\
  \bf \  Jiazhen Pan, 
  \ Ayush Noori,
  \ Sean Wu,
  \ Honghan Wu, \\
  \bf  \ Fenglin Liu\thanks{Corresponding authors.},
  \ David A. Clifton\footnotemark[2] 
  \\
  \textsuperscript{1}University of Oxford \ \ \
  \textsuperscript{2}University College London \ \ \
  \textsuperscript{3}Technical University of Munich \\
  \textsuperscript{4}Oxford-Suzhou Centre for Advanced Research, China \\
  {\tt jinge.wu.20@ucl.ac.uk, fenglin.liu@eng.ox.ac.uk} 
  \\ \\
}

\begin{document}

\maketitle

\begin{abstract}

Reproducing and comparing deep research agents today is hard: the same backbone evaluated on the same benchmark can report different accuracies across papers because the harness and tool registry differ, and integrating a new model into a comparable evaluation surface costs weeks of model-specific engineering. These are symptoms of a broader reproducibility problem in deep research agent research. Here, we introduce BioMedArena, an open-source toolkit that addresses this reproducibility gap and provides an arena for comparing deep research agents under a shared evaluation environment.
BioMedArena decouples six layers of biomedical agent evaluation—benchmark loading, tool exposure, tool selection, harness mode, context management, and scoring—and exposes 166 biomedical benchmarks and 75 biomedical tools across 9 functional families.
Adding a new model, benchmark, or tool can be accomplished with a few-line provider adapter. Beyond evaluation infrastructure, BioMedArena ships a library of high-quality reference components---6 agent harnesses (including our proposed \textsc{Mutual-Evolve}) and 6 context-management strategies---any of which can be equipped on any backbone. Equipping these components substantially improves all 12 backbones; on each of 8 representative biomedical benchmarks, the best equipped backbone surpasses prior state-of-the-art (SOTA), by +15.01 percentage points on average \footnote{The toolkit, configurations, and per-task traces are available at \url{https://github.com/AI-in-Health/BioMedArena}.}.

\vspace{-22pt}
\end{abstract}

\begin{figure}[H]
  \centering
  \includegraphics[width=0.75\textwidth]{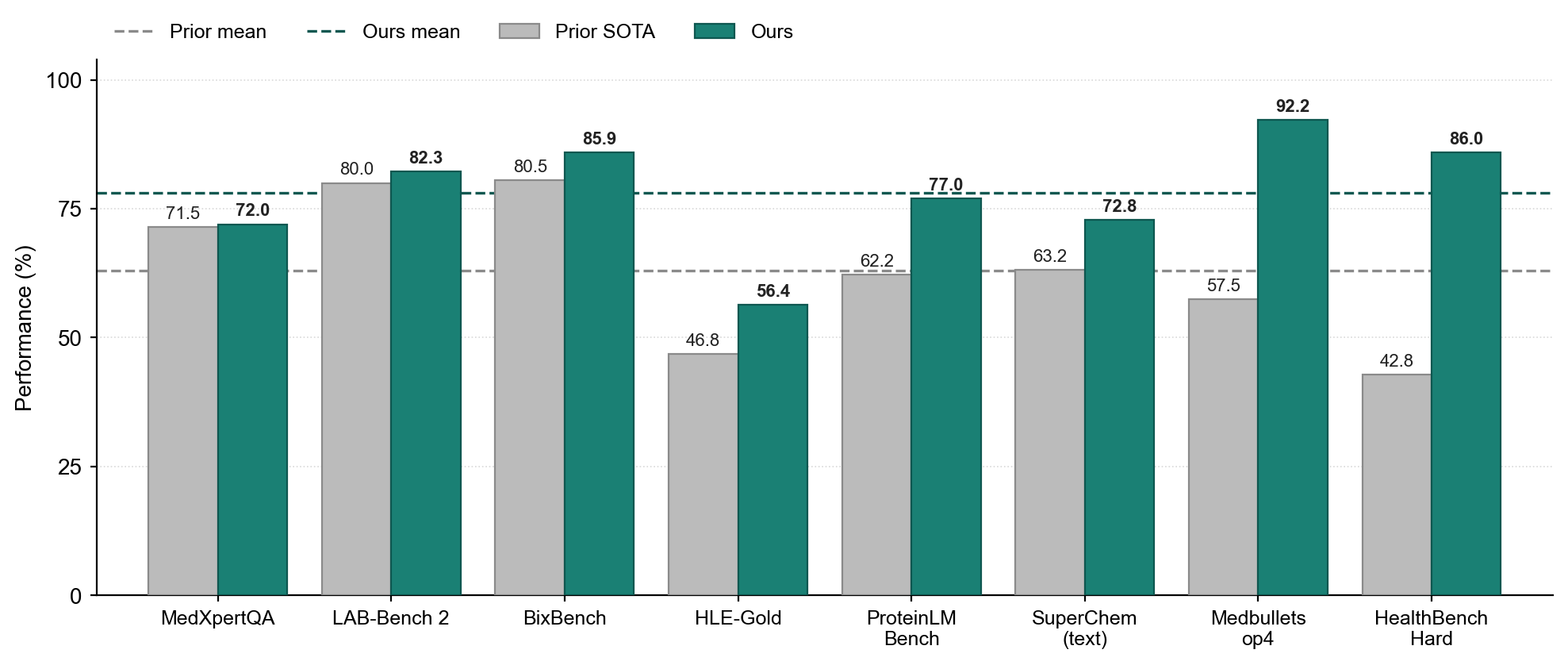}
\vspace{-12pt}
  \caption{Performance of deep research agents under BioMedArena across 8 representative biomedical benchmarks. For each benchmark, we plot the best BioMedArena-equipped backbone (\emph{Ours}) against the prior published SOTA.}
  \label{fig:overview}
  \vspace{0pt}
\end{figure}

\section{Introduction}\label{sec:intro}
\vspace{-5pt}
Deep research agents leverage tool use, long-horizon planning, and multi-agent orchestration to tackle complex research tasks and advance human knowledge~\citep{deepresearch_survey2025}. They have recently emerged as a focal point of LLM research, with benchmarks such as Humanity's Last Exam (HLE)~\citep{hle2025} drawing widespread attention as a proving ground for the research capability of LLMs with tools. However, even when evaluated on the same benchmarks with the same backbone model, reported performance metrics diverge across papers or require substantial engineering effort to reproduce~\citep{kapoor2025holistic}.
Most existing studies do not open-source their evaluation harness and tool implementations, resulting in a reproducibility crisis in deep research agent research.
For example, a researcher who wants to evaluate whether Claude Opus 4.6~\citep{claude_opus_46} or Gemini 3.1 Pro~\citep{gemini31_pro} is the stronger backbone for agentic literature retrieval grounded in PubMed, or whether a newly released open-weight protein model is competitive with closed-weight backbones on variant interpretation, must evaluate the candidate systems under the same tool registry, same iteration cap, or harness strategies. Assembling this evaluation surface takes days to weeks of model-specific engineering.
This is an agent-specific instance of the reproducibility problem: an engineering cost that deep research agent researchers pay separately, repeatedly, and rarely amortize across papers.
This cost arises along 3 concrete axes: \emph{Harnesses are not comparable}: The agent harness is one of the most important components for enabling LLMs to perform deep research. However, existing deep research agent systems each implement their own loop (e.g., ReAct-style~\citep{react2023}, plan-execute, self-consistency (majority voting)~\citep{wang2022self}, or dialog-turn~\citep{autogen2023,crewai2023}), and the same backbone can produce substantially different performance across them.
\emph{Tool registries are not shared}: each system~\citep{biomni2025,medagentbench2024,agentclinic2024} bundles its own typed tools with its own schema, and a tool implemented once is not shared or reusable for different models.
\emph{Scoring is not
unified}: deterministic match,
LLM-as-judge~\citep{llmjudge2023}, and code-execution paradigms
are mixed inconsistently, and the choice of judge model itself
shifts reported accuracy. The downstream consequence is that
adding a new model or agent to an evaluation task requires re-implementing model-specific code in each of
these 3 axes separately. There is no shared environment under which different deep research agents can be compared head-to-head.

To address this gap, we introduce BioMedArena, an open-source toolkit that decouples 6 layers of biomedical deep research agent evaluation --- benchmark loading, tool exposure, tool selection, harness mode, context management, and scoring --- and exposes 166 biomedical benchmarks, 75 biomedical tools across 9 functional families, and a benchmark-aware scoring router. 
With BioMedArena, evaluating a new model as a deep research agent on over a hundred evaluation datasets requires no benchmark-specific engineering, no tool-integration work, and no scoring logic on the part of the model developer; rather, a small, \mbox{often < 100-line} provider adapter is sufficient for comparison with previous agents in deep research capability under the same evaluation environment and settings.

BioMedArena also implements 6 agent harnesses and 6 context management strategies, any of which can be equipped on any backbone for comparable evaluation against every other agent.
Lastly, we present \textsc{Mutual-Evolve}, a deep research agent harness in which multiple parallel solvers share key intermediate findings through a typed Global Workspace of four banks (errors, skills, tools, and guides), preserving majority-vote self-consistency while letting solvers share what they learn and aggregate answers by the evidence behind them, not only the final choice.

We validate BioMedArena on 12 backbones over 8 benchmarks run under the same evaluation environment, in which the harness, tool registry, and scoring policy are fixed for fair comparison.
The 12 backbones include 5 open-source and 7 closed-source models, and the 8 benchmarks cover medicine, biology, chemistry, genomics, laboratory research, and multidisciplinary reasoning. To our knowledge, this is the first regularly comparable biomedical agent evaluation at this scale.
Beyond serving as an evaluation harness, BioMedArena's reference components---its agent harnesses, tool registry, and context-management strategies---are themselves strong: as Figure~\ref{fig:overview} shows, equipping backbones with them substantially improves their research capability, and on each of the 8 benchmarks the best equipped backbone surpasses the prior best result, by +15.01 percentage points on average.

\begin{figure}[t]
  \centering
  \includegraphics[width=\textwidth]{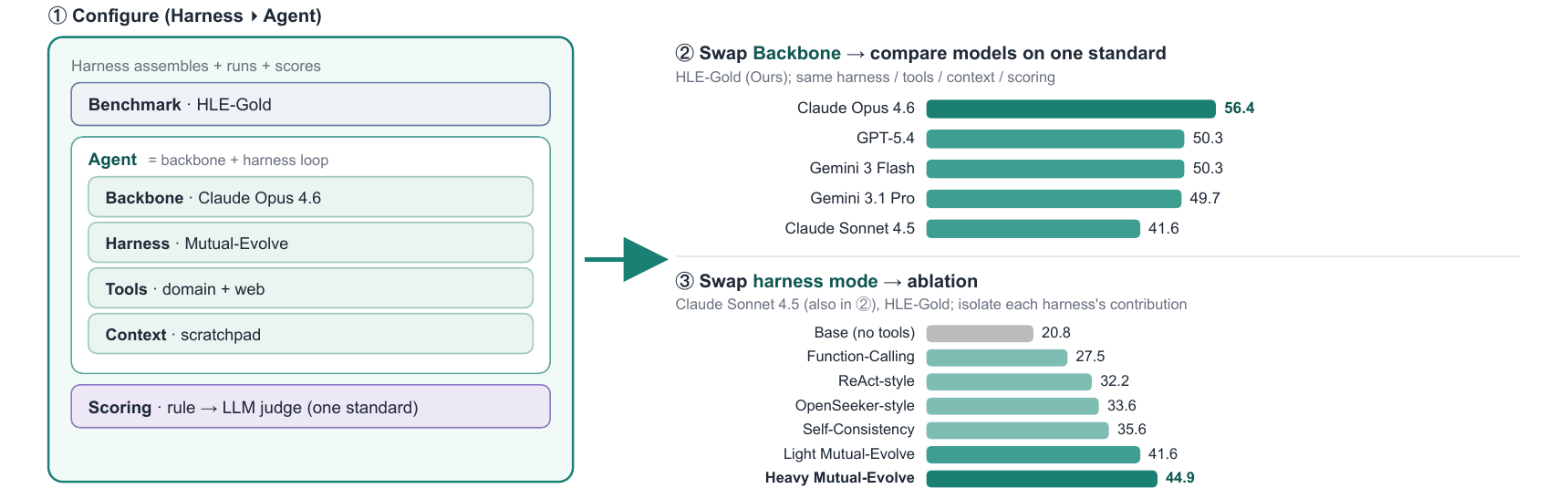}
  \vspace{-5pt}
  \caption{BioMedArena casts evaluation as a configuration: the harness runs an \emph{agent}---a backbone driving a harness loop over a benchmark-specific tool subset with a context manager---and scores it through one scoring router (left). Since every layer is swappable, sweeping one axis under fixed scoring isolates its effect: swapping the \textbf{backbone} compares models (top right); swapping the \textbf{harness mode} ablates each harness (bottom right). }
  \label{fig:toolkit_overview}
\end{figure}

\vspace{-5pt}
\section{Related Work}\label{sec:related}
\vspace{-5pt}
The biomedical deep research agent landscape today consists of three kinds of artefacts, none of which is an integrated evaluation toolkit.

\emph{Single-system agents} such as Biomni~\citep{biomni2025}, MedAgentBench~\citep{medagentbench2024}, and AgentClinic~\citep{agentclinic2024} each ship one agent or one environment; LAB-Bench, LAB-Bench~2, and BixBench~\citep{labbench2024,labbench2_2025,bixbench2025} similarly each target one biomedical capability. None of them offers a substrate under which different agents or harnesses can be compared head-to-head across many biomedical tasks.

\emph{Static evaluation suites} such as \texttt{lm-eval-harness}~\citep{lmevalharness}, HELM~\citep{helm2023}, and MedHELM~\citep{medhelm2025} scale benchmark coverage but only under single-call prompting; agentic execution, tool use, and context management are out of scope. Benchmarks built specifically for deep research agents, such as DeepResearch Bench~\citep{deepresearchbench2025}, do evaluate agentic execution, but as fixed, general-domain benchmarks they assess a single agent configuration rather than offering a reconfigurable harness, tool, and scoring environment for head-to-head comparison.

\emph{General-purpose agent frameworks} such as Inspect~\citep{inspect2024} support tool use but are not biomedically specialised and leave benchmark, tool, and scoring wiring to the user.

BioMedArena fills the gap between these three: it is an open-source toolkit that provides a shared and reproducible evaluation environment for biomedical deep research agents, unifying 166 benchmarks under one task interface, exposing a typed biomedical tool registry, letting the user switch among 6 agent harnesses and 6 context-management strategies without modifying the benchmark, and routing scoring through a benchmark-aware judge. We build on agent-architecture primitives, including iterative tool-use loops, hierarchical delegation, and summarization-based compression~\citep{react2023,autogen2023,crewai2023,metagpt2023,multiagentdebate2023}; our contribution is the integration layer that turns biomedical deep research agent evaluation from a repeated, system-specific engineering effort into shared, head-to-head infrastructure. \autoref{tab:related} of Appendix contrasts the role each system plays in the landscape and the features each one supports out of the box.

\section{BioMedArena}\label{sec:toolkit}

As shown in Figure~\ref{fig:toolkit_overview}, BioMedArena is an extensible toolkit rather than a fixed benchmark script. The design decouples benchmark loading, tool exposure, harness mode, context management, model backend, and scoring, so that any one of these can be replaced, ablated, or extended without touching the others. This decoupling is what reduces this recurring engineering cost to a single provider adapter: once a new model is registered, every benchmark, every tool, every harness mode, and every scoring policy is immediately available to it. This section introduces the toolkit in detail.

\paragraph{A unified biomedical task interface.} BioMedArena exposes 166 biomedical benchmarks, including LAB-Bench\,2 \citep{labbench2_2025}, BixBench \citep{bixbench2025}, MedXpertQA~\citep{medxpertqa2024},  SuperChem~\citep{superchem2025}, and HLE \citep{hle2025}, the widely used benchmarks in existing deep research agent works.
Each benchmark
is normalized into a common task object containing question
text, expected answer, answer type, scoring metadata, and
benchmark-specific context fields. The normalization is programmatic, not per-sample manual: each benchmark is declared by a one-line specification (source repository, task type, and domain), and a rule-driven loader maps raw dataset fields into the common schema via field-name heuristics dispatched by task type, with optional per-field overrides in the same specification; a second rule-based layer canonicalizes answers and text for scoring. This normalization is the
bridge between heterogeneous biomedical evaluation sources and a
single agentic execution interface: subsequent harnesses,
tool selection, scoring, and trace logging do not need to know
what the original benchmark format was. Benchmarks span
medicine, biology, genomics and bioinformatics, pathology,
chemistry, and multi-discipline reasoning.

\paragraph{A tool registry.} 
BioMedArena registers 75 tools across 9 functional families: literature and search, clinical reference and decision support, genomics and transcriptomics, proteins and structure, chemistry and biochemistry, disease biology, variants and pathways and ontology, imaging, and code-statistics-survival (Appendix Figure~\ref{fig:tool_overview}). Each tool carries a name, a structured parameter signature, a return type, and a category tag, which lets the harness execute, trace, and analyze tool invocations uniformly across benchmarks and providers. New tools can enter the registry by simply adding a schema-and-handler pair and are immediately available to every benchmark.

\paragraph{Agent harness.} 

The six harnesses cover the most common deep-research setups:
(i) \textsc{Function-Calling}, a single forward pass with one round of tool use, serving as a single-step tool-augmented baseline;
(ii) \textsc{ReAct-style}~\citep{react2023}, a Thought-Action-Observation loop in which the model alternates between reasoning steps and tool calls until it emits a final answer;
(iii) \textsc{OpenSeeker-style}~\citep{du2026openseeker}, a search-oriented harness that decomposes the task into an explicit sub-query plan and iteratively refines its retrieval trajectory before synthesizing the final answer;
(iv) \textsc{Self Consistency}~\citep{wang2022self}, a sampling-based harness that runs $N$ independent agent trajectories on the same task and selects the final answer by majority vote across rollouts, trading additional inference cost for robustness to single-trajectory failures;
(v) \textsc{Light Mutual-Evolve}, our proposed Mutual-Evolve agent loop over the benchmark-aware tool subset with thinking off, designed as the lightweight default for typical biomedical tasks; and
(vi) \textsc{Heavy Mutual-Evolve}, the same Mutual-Evolve loop with thinking on, the full tool registry available, and a minimum-iteration floor that forces the agent to perform deep research before answering.
We will detail the proposed mutual-evolve harness in Section~\ref{sec:mutual_evolve}.

\paragraph{Context management.} 
Context management determines what enters the model's input during tool iterations and reasoning. Long biomedical traces accumulate hundreds of thousands of tokens of tool output, model reasoning, and tool-call records before a final answer is produced, potentially causing context overflow. Therefore, context management is one of the key design factors in whether the agent succeeds or fails.
As shown in Table~\ref{tab:context-strategies} of the Appendix, BioMedArena implements comprehensive context-management strategies: planning, summarization, clearing, truncation, memory, and rollback.
The six strategies are independently togglable and compose freely: \emph{planning} maintains compact working notes of intermediate findings, unresolved subgoals, and accumulated evidence across multi-step tool use; \emph{memory} persists salient facts and prior findings to an external store and re-injects relevant entries on demand; \emph{summarization} compresses earlier dialogue and tool outputs into shorter summaries while preserving recent context once the trace exceeds a length threshold; \emph{clearing} replaces stale or verbose payloads, such as old tool outputs, reasoning traces, or large media, with compact placeholders; \emph{truncation} retains task-relevant context through sliding windows, first-last retention, or token budgets; and \emph{rollback} removes the most recent low-value assistant or tool turn and inserts corrective guidance to break out of repeated queries, tool errors, or early loop formation.

\paragraph{Scoring router.} Different benchmarks require different scoring policies, so the toolkit implements a scoring router. Following common practice, for open-ended and multiple-choice tasks, the LLM judge is the primary scorer, with the deterministic scorer recording metadata. For structured tasks (including exact match, numeric comparison, and regex), if the model follows the instruction to output a structured answer that can be extracted, a deterministic scorer runs first. If a structured answer cannot be extracted or the deterministic scorer returns ``incorrect,'' but the model produced a non-empty answer, an LLM judge is invoked to catch semantically correct answers that fail strict matching—near-equal numeric values (``0.545'' vs.~``0.55''), formatting equivalents (``1/3'' vs.~``1:3''), and similar cases.

All experiments in Section~\ref{sec:experiments} use a single fixed judge model regardless of the model and benchmarks used for evaluation, which prevents scorer-induced variance.

\begin{figure}[t]
  \centering
  \includegraphics[width=0.8\textwidth]{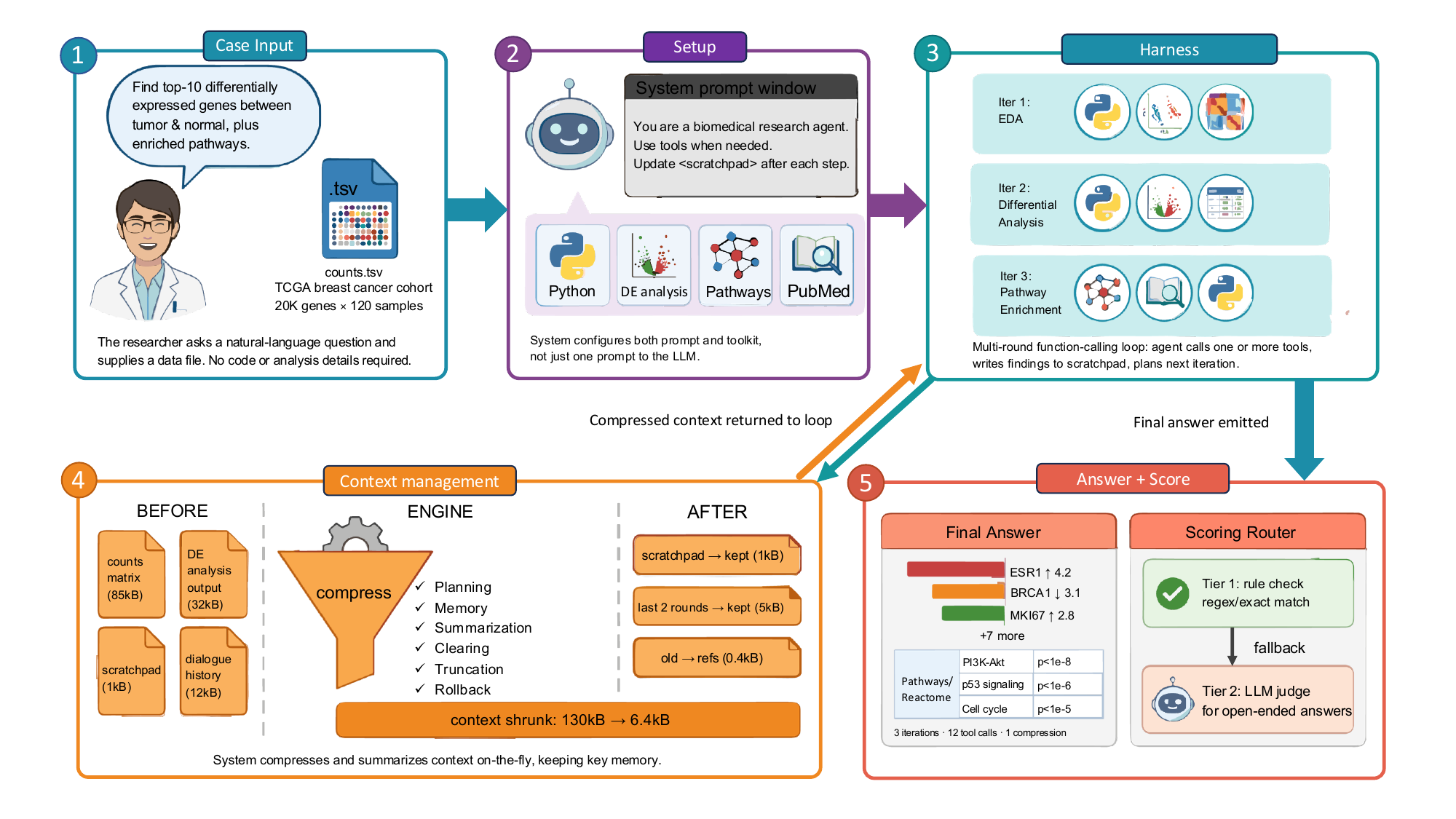}
  \vspace{-10pt}
  \caption{Dataflow of a biomedical deep research agent in BioMedArena. A natural-language question with an attached data file (1) is dispatched through a system that configures both prompt and toolkit (2) into a multi-round agent loop (3) that issues tool calls, writes findings to a scratchpad, and plans the next iteration. Long traces are compressed on-the-fly by the context-management engine (4), which composes planning, memory, summarization, clearing, truncation, and rollback to keep key state while shrinking bulk payloads. The final answer is routed through a two-tier scoring layer (5)---a deterministic rule check followed by an LLM judge for open-ended responses.}
  \label{fig:data_flow}
\end{figure}

\vspace{-5pt}
\section{Mutual-Evolve Harness}
\label{sec:mutual_evolve}
\vspace{-5pt}

Figure~\ref{fig:data_flow} shows the dataflow of a biomedical agent in BioMedArena: at each iteration, it issues a model call with the current message stack and tool subset, parses any returned tool calls, executes them in parallel where independent, and appends the structured results before the next iteration. As an important component of the deep research agent architecture, the agent harness is introduced here to guide the model through deeper, multi-turn interactions with tools to produce more accurate answers.

LLM agents for closed-domain reasoning typically use either single-rollout execution or self-consistency over $N$ independent rollouts~\citep{wang2022self}. A single rollout is bounded by the stability of one trajectory, so an early misstep often propagates to a wrong final answer. Self-consistency reduces this variance via majority voting, but two problems remain. First, rollouts are \emph{informationally isolated}: a useful intermediate finding in one rollout---a relevant guideline retrieved, a calculator output verified, a misleading hypothesis ruled out---does not help the others. Second, voting acts \emph{only on final answers}, so a minority rollout that found the correct evidence is silently overridden by a majority that agrees on a plausible but wrong one.

A natural fix is to let rollouts share intermediate findings through a common workspace, but naive sharing can be worse than no sharing. If solvers exchange partial reasoning from the start, an early speculative claim anchors the others and collapses ensemble diversity before alternatives emerge. If every intermediate thought is broadcast indiscriminately, confident guesses become indistinguishable from verified facts, reinforcing rather than correcting collective error.

\begin{figure}[t]
  \centering
  \includegraphics[width=1\textwidth]{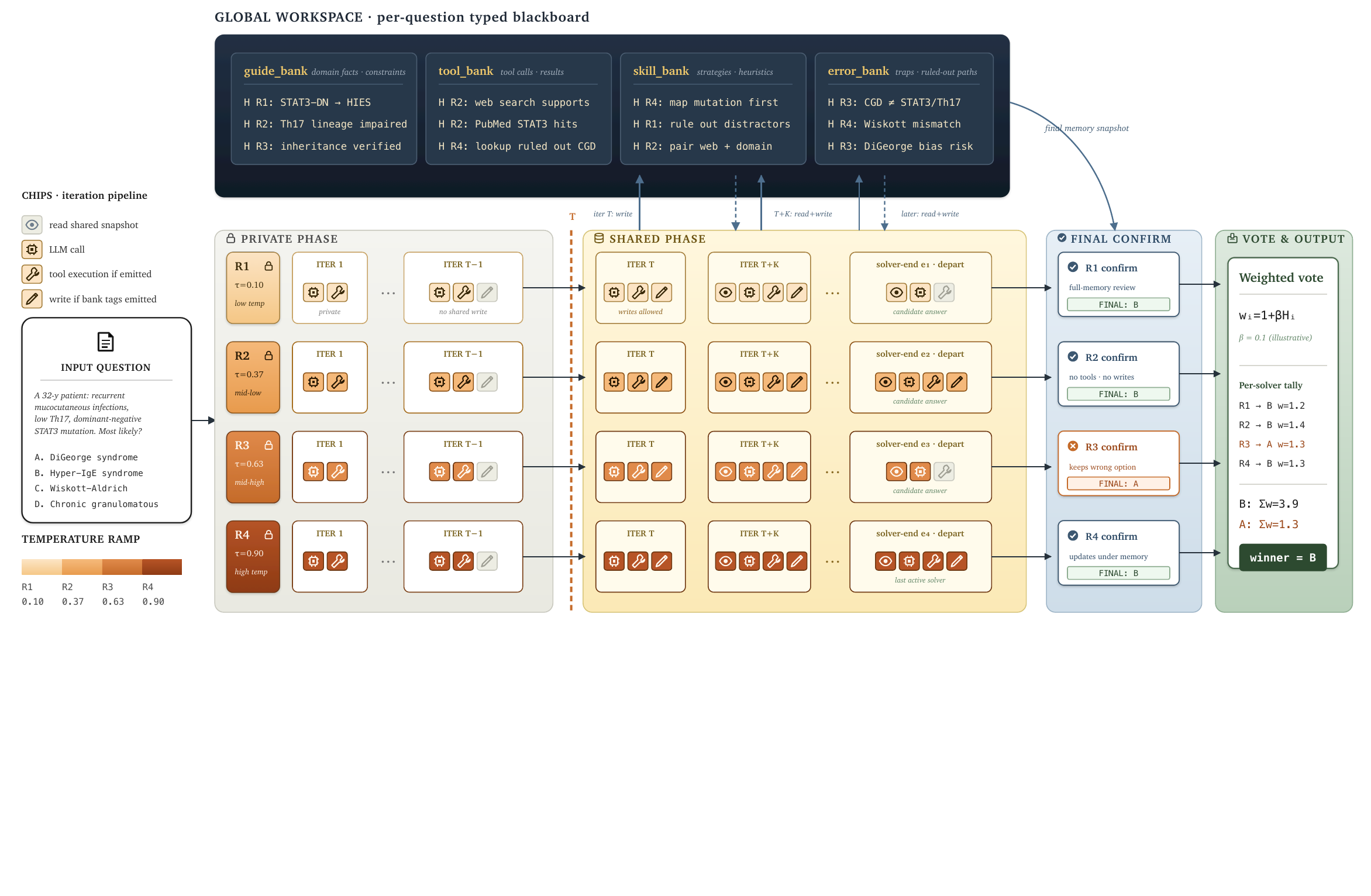}
  \vspace{-80pt}
\caption{\textsc{Mutual-Evolve} workflow. For each question, $N$ parallel solvers at distinct temperatures first explore privately, then share findings through a Global Workspace at iteration $T$. The workspace has four typed banks (guide, tool, skill, error); solvers read it every $K$ iterations and may terminate at different end iterations $e_i$. Once all solvers finish, each performs a text-only final confirmation over the full workspace, and answers are aggregated by contribution-weighted voting. The framework-level procedure is given in Algorithm~\ref{alg:mutual_evolve} of our Appendix.}
\label{fig:harness_overview}
\vspace{-5pt}
\end{figure}

\textsc{Mutual-Evolve} resolves this tension with three design principles, each realized by a component in Figure~\ref{fig:harness_overview}. {Private exploration} requires each solver to reason in isolation for a fixed number of initial iterations, with temperatures spaced across the cohort to ensure trajectory diversity. {Selective typed sharing} then opens a Global Workspace with four banks---errors, skills, tools, and guides---that keep contributions distinguishable by epistemic role. {Contribution-weighted voting} aggregates final answers, giving more weight to solvers whose findings entered the shared record.

\vspace{-5pt}
\subsection{Framework Overview}
\label{sec:me_overview}
\vspace{-5pt}

\textsc{Mutual-Evolve} solves a single question with a cohort of $N$ parallel solvers that share a per-question Global Workspace, instantiated empty for every question and discarded once it is resolved. Each solver uses the same task prompt and tool inventory, differing only in sampling temperature.
Execution proceeds in synchronized rounds. Within each round, every active solver reads from the workspace if scheduled, issues one LLM call that may emit workspace writes and tool calls, executes any requested tools, and waits at a barrier until the remaining solvers finish the round. Rounds are partitioned into a {private phase} of $T$ rounds, during which the workspace is unreachable, and a {shared phase} beginning at round $T$, in which writes are permitted, and reads occur every $K$ rounds starting from round $T+K$. A solver may commit a candidate answer once it has accumulated at least $L_{\min}$ tool-using iterations. After all solvers commit or terminate, a {Final Confirmation} step injects the complete workspace back into each completed solver for a text-only review, and a {contribution-weighted vote} over the confirmed answers produces the final prediction.
The framework's hyperparameters are the cohort size (number of solvers) $N$, the temperature schedule $\{\tau_1, \ldots, \tau_N\}$, the private-phase length $T$, the read interval $K$, the minimum tool-use depth $L_{\min}$, and the voting coefficient $\beta$.

\vspace{-5pt}
\subsection{Parallel Solvers and Diversity}
\label{sec:me_solvers}
\vspace{-5pt}

Each of the $N$ solvers maintains a private conversational state, so the only channel through which solvers can influence one another is the Global Workspace. Each solver $i$ samples at a distinct temperature $\tau_i$ from a fixed schedule. Because the private phase forbids inter-solver communication, trajectory diversity during the first $T$ rounds depends entirely on the variance of the sampling distribution, and identical temperatures would yield highly correlated rollouts. In our implementation, we space $\tau_i$ evenly over $[0.1, 0.9]$, but the schedule itself is a hyperparameter.
At each round, a solver reads any scheduled workspace snapshot, issues one LLM call that may produce reasoning text, bank-tagged workspace writes (shared phase only), and tool-call requests, executes the requested tools, and waits at the round barrier.

\begin{table*}[t]
\centering
\scriptsize
\caption{
Ablation study of our design. We adopt Claude Sonnet 4.5~\citep{claude_sonnet_45} as the backbone and report accuracy (\%) results on HLE-Gold and BixBench benchmarks. 
}
\label{tab:ablation}
\setlength{\tabcolsep}{1pt} 
\begin{tabular}{@{}lccccccccccccc@{}}
\toprule
 \multirow{2}{*}[-3pt]{{Harness}} &  \multicolumn{6}{c}{{Context Management}} & \multirow{2}{*}[-3pt]{\begin{tabular}[c]{@{}c@{}} Solver \\ $N$ \end{tabular}}  & \multirow{2}{*}[-3pt]{\begin{tabular}[c]{@{}c@{}}  Private \\  Phase $T$ \end{tabular} }  
 & \multirow{2}{*}[-3pt]{\begin{tabular}[c]{@{}c@{}}  Shared \\  Phase $K$ \end{tabular} }  
&  \multirow{2}{*}[-3pt]{ \begin{tabular}[c]{@{}c@{}}  Final \\  Confirm\end{tabular}  } 
 &  \multirow{2}{*}[-3pt]{ \begin{tabular}[c]{@{}c@{}}  Weighted \\  Voting \end{tabular}  }  & \multirow{2}{*}[-3pt]{HLE-Gold} & \multirow{2}{*}[-3pt]{BixBench}
\\ \cmidrule(lr){2-7} 

&
 \begin{tabular}[c]{@{}c@{}} Plan
\end{tabular}   &  \begin{tabular}[c]{@{}c@{}} Summ.
\end{tabular}     & \begin{tabular}[c]{@{}c@{}} Clear\end{tabular} & \begin{tabular}[c]{@{}c@{}} Trunc. \end{tabular} & \begin{tabular}[c]{@{}c@{}} Memo. \end{tabular} & \begin{tabular}[c]{@{}c@{}} Rollb. \end{tabular} 

\\ 
\midrule

Base & -  & - & - & -  & -  & -  & -  & -  & -  & -  & -  & 20.8  & 17.1

 \\

\midrule

 Function-Calling   & -   & - & - & -  & -  & -  & -  & -  & -  & - & -  & 27.5 & 28.3
\\ 
ReAct-style & $\checkmark$ & $\checkmark$  & - & - & -  & -  & -  & -  & -  & -  & -  & 32.2  & 34.1
\\
OpenSeeker-style   & $\checkmark$ & $\checkmark$  & - & - & -  & -  & -  & -  & -  & -  & - & 33.6 & 33.1   \\
Self-Consistency  & $\checkmark$ & $\checkmark$  & - & - & -  & -  & -  & -  & -  & -  & -  & 35.6 & 35.6 \\
Light Mutual-Evolve  & $\checkmark$ & $\checkmark$  & - & - & -  & -  & 4  & 10  & 3  &  $\checkmark$ & $\checkmark$ & 41.6 & 42.4 \\
Heavy Mutual-Evolve  & $\checkmark$ & $\checkmark$   & - & - & -  & -  & 4  & 10  & 3  &  $\checkmark$ & $\checkmark$  & \bf 44.9 & \bf 43.4\\

\midrule  
 \multirow{14}{*}[-0pt]{Light Mutual-Evolve} & $\checkmark$ & $\checkmark$  & $\checkmark$  & - & -  & -  & 4  & 10  & 3  &  $\checkmark$ & $\checkmark$  & 36.9 & 37.1\\
  & $\checkmark$ & $\checkmark$  & - & $\checkmark$  & -  & -  & 4  & 10  & 3  &  $\checkmark$ & $\checkmark$  &34.9  & 40.0\\
  & $\checkmark$ & $\checkmark$  & - & - & $\checkmark$   & -  & 4  & 10  & 3  &  $\checkmark$ & $\checkmark$  & 39.6 & 39.0\\
    & $\checkmark$ & $\checkmark$  & - & - & -   & $\checkmark$  & 4  & 10  & 3  &  $\checkmark$ & $\checkmark$   & 40.3 & 41.5\\
\cmidrule(lr){2-14}

  & $\checkmark$ & $\checkmark$   & - & - & -  & -  & 2  & 10  & 3  &  $\checkmark$ & $\checkmark$  & 38.9 &  38.5  \\
& $\checkmark$ & $\checkmark$   & - & - & -  & -  & 8 & 10  & 3  &  $\checkmark$ & $\checkmark$ & 36.9 & 40.5\\ \cmidrule(lr){2-14}
& $\checkmark$ & $\checkmark$   & - & - & -  & -  & 4  & -  & 3  &  $\checkmark$ & $\checkmark$  & 32.9 & 32.2\\
& $\checkmark$ & $\checkmark$   & - & - & -  & -  & 4  & 5  & 3  &  $\checkmark$ & $\checkmark$  & 37.6 & 36.1\\
& $\checkmark$ & $\checkmark$   & - & - & -  & -  & 4  & 15  & 3  &  $\checkmark$ & $\checkmark$  & 39.6 & 37.6\\ \cmidrule(lr){2-14}

& $\checkmark$ & $\checkmark$   & - & - & -  & -  & 4  & 10  & 1  &  $\checkmark$ & $\checkmark$ & 38.3 & 38.0
\\
& $\checkmark$ & $\checkmark$   & - & - & -  & -  & 4  & 10  & 5  &  $\checkmark$ & $\checkmark$ & 40.3   & 35.1
\\ \cmidrule(lr){2-14}
& $\checkmark$ & $\checkmark$   & - & - & -  & -  & 4  & 10  & 3  &  -  & $\checkmark$  & 37.6 & 38.5
\\
& $\checkmark$ & $\checkmark$   & - & - & -  & -  & 4  & 10  & 3  &  $\checkmark$  & - & 40.9 & 40.4
\\
\bottomrule
\end{tabular}
\end{table*}

\textbf{Typed Global Workspace.}
The Global Workspace is organized into four banks of distinct epistemic kinds. The {error bank} records failed reasoning paths and ruled-out hypotheses; the {skill bank} records reusable strategies and reasoning or tool-use heuristics; the {tool bank} records noteworthy tool invocations and their results; and the {guide bank} records domain facts, constraints, and key pieces of evidence. The four-way typing is intentional: it discourages undifferentiated dumping of internal monologue and lets a reader of the workspace quickly locate the kind of information it currently needs. Writes are model-initiated. A solver contributes an entry by emitting one of the bank tags---\texttt{<error\_bank>}, \texttt{<skill\_bank>}, \texttt{<tool\_bank>}, or \texttt{<guide\_bank>}---within its LLM output during the shared phase; a solver that emits no bank tags in a round writes nothing. Whether to share, and what to share, is left to the model. When a solver reads the workspace, it receives a formatted snapshot of all non-empty banks at that moment, with each entry annotated by provenance.

\textbf{Private phase.}
During the first $T$ rounds, the Global Workspace is unreachable. This phase lets the temperature schedule translate into trajectory diversity: setting $T$ too small gives diversity no time to develop, while setting it too large wastes effort on isolated reasoning when collaborative refinement could have begun.

\textbf{Shared phase.}
From round $T$ onward, writes are permitted at every round, while reads occur at the start of every $K$-th round through a formatted snapshot injected into the solver's context. Setting $K=1$ makes shared findings immediately visible in the next round; larger $K$ amortizes context cost at the price of staler reads.

\textbf{Synchronization and departure.}
Rounds are synchronized by a barrier: after completing round $t$, each active solver waits until the rest finish. A solver that commits a candidate answer or terminates abnormally departs at its {end iteration} $e_i$ and no longer blocks the cohort; $e_i$ generally differ across solvers, since each decides independently when to commit.

\textbf{Final confirmation.}
Once every solver has terminated, each completed solver receives a snapshot of the entire Global Workspace injected into its conversational history, and is prompted to review its candidate answer against the consolidated evidence and emit a final response ending with \texttt{FINAL\_ANSWER: <answer>}. This step is text-only and produces no further workspace writes. Its role is twofold: it integrates information across the cohort---exposing each solver to findings it may not have read in time---and equalizes the information available before voting.

\textbf{Contribution-weighted voting.}
Each solver $i$ casts a vote for its extracted answer with weight:
\begin{equation}
w_i \;=\; 1 + \beta\, H_i,
\label{eq:me_weight} \nonumber
\end{equation}
where $H_i$ is the number of entries that solver $i$ contributed to the Global Workspace and $\beta \ge 0$. The base weight ensures every completed solver retains a vote, and the additive term gives greater influence to solver whose findings entered the collaborative record. The final prediction is $\hat{a} = \arg\max_a \sum_{i:\, a_i = a} w_i$, with ties broken in favor of the answer that first appears in the cohort's response order.

\begin{table}[t]
  \caption{Accuracy (\%) of 12 backbones evaluated as deep research agents on 8 biomedical benchmarks. Each backbone occupies 2 sub-rows: \textsc{Baseline}, the backbone answering from extended thinking alone with no tools, and \textsc{Ours}, the same backbone equipped with BioMedArena's default \textsc{Light Mutual-Evolve} harness. Bold marks the best result in each benchmark column.}
  \label{tab:main_results}
  \centering
  \scriptsize
  \setlength{\tabcolsep}{3pt}
  \renewcommand{\arraystretch}{1.05}
  \resizebox{\textwidth}{!}{%
  \begin{tabular}{c l c cccc ccc c ccc c}
    \toprule
    \multirow{2}{*}{\textbf{Types}} & \multirow{2}{*}{\textbf{Backbones}} & \multirow{2}{*}{\textbf{Settings}}
       & \shortstack{HealthBench \\ Hard}
       & MedXpertQA
       & \shortstack{ProteinLM \\ Bench}
       & Medbullets
       & \multicolumn{3}{c}{SuperChem (500)}
       & BixBench
       & \multicolumn{3}{c}{HLE-Gold (149)}
       & LAB-Bench\,2 \\
    \cmidrule(lr){8-10}   \cmidrule(lr){12-14}
       & & & (1000)
       & (2450)
       & (944)
       & (308)
       & Text & Image & Total\textsuperscript{$\dagger$}
       & (205)
       & Bio & Chem & Total
       & (821) \\
    \midrule
    & SOTA & --- & 42.8\textsuperscript{~\citep{healthbench_hard_sota_muse}} & 71.5\textsuperscript{~\citep{medxpertqa_sota_gemini}} & 62.2\textsuperscript{~\citep{internlm2_protein}} & 57.5\textsuperscript{~\citep{medreason8b}} & 63.2\textsuperscript{~\citep{mirothinker2026}} & --   & 38.5\textsuperscript{~\citep{superchem2025}} & 80.5\textsuperscript{~\citep{gpt55_2026}} & 44.2\textsuperscript{~\citep{edison2026}} & 53.2\textsuperscript{~\citep{edison2026}} & 46.8\textsuperscript{~\citep{edison2026}} & 80.0\textsuperscript{~\citep{edison2026}} \\
    \midrule
    \multirow{10}{*}{\rotatebox[origin=c]{90}{\textbf{Public LLMs}}}
      & \multirow{2}{*}{Trinity-Large-Thinking~\citep{trinity2025}}
        & Baseline & 41.9   & 33.3 & 39.0 & 77.3 & 23.8 & --   & --   &  5.4 & 15.0 &  7.1 & 12.8 & 43.5 \\
      & & Ours     & 47.7   & 44.2 & 69.5 & 82.1 & 30.6 & --   & --   & 26.3 & 19.6 & 19.0 & 19.5 & 61.0 \\
      \cmidrule(l){3-15}
      & \multirow{2}{*}{Nemotron-3-Super-120B-A12B~\citep{nemotron3_2025}}
        & Baseline & 58.7   & 37.8 & 50.5 & 79.2   & 32.5   & --   & --   & 13.2   & 13.1 &  4.8 & 10.7 & 43.6 \\
      & & Ours     & 65.4   & 45.9 & 63.1 & 82.5   & 39.6   & --   & --   & 10.7   & 28.0 & 33.3 & 29.5 & 62.6 \\
      \cmidrule(l){3-15}
      & \multirow{2}{*}{INTELLECT-3.1~\citep{intellect31_2026}}
        & Baseline & 41.1   & 36.2 & 43.5 & 72.7 & 27.9 & --   & --   &  5.4 & 20.6 &  9.5 & 17.4 & 43.5 \\
      & & Ours     & 48.2   & 37.5 & 65.6 & 75.0 & 29.4 & --   & --   & 20.0 & 26.2 & 19.0 & 24.2 & 57.9 \\
      \cmidrule(l){3-15}
      & \multirow{2}{*}{GLM-4.5~\citep{glm45_2025}}
        & Baseline & 42.0 & 35.3 & 59.2 & 79.2   & 21.1   & --   & --   & 23.9 & 16.8 &  2.4 & 12.8 & 44.8 \\
      & & Ours     & 45.3 & 36.5 & 63.1 & 81.2   & 28.3   & --   & --   & 30.2 & 28.0 & 28.6 & 28.2 & 61.5 \\
      \cmidrule(l){3-15}
      & \multirow{2}{*}{Qwen3-235B-A22B~\citep{qwen35_2026}}
        & Baseline & 27.9   & 40.5 & 54.0 & 79.5 & 23.4 & --   & --   & 22.4 & 15.0 & 14.3 & 14.8 & 42.5 \\
      & & Ours     & 32.4   & 40.8 & 62.2 & 83.1 & 37.7 & --   & --   & 35.1 & 28.0 & 14.3 & 24.2 & 57.9 \\
    \midrule
    \multirow{14}{*}{\rotatebox[origin=c]{90}{\textbf{Commercial  \ \ LLMs}}}
      & \multirow{2}{*}{GPT-5.4~\citep{gpt54_2025}}
        & Baseline & 61.3 & 44.7 & 67.6 & 84.8 & 57.7 & 22.6 & 41.2 & 40.0 & 43.0 & 33.3 & 40.3 & 48.5 \\
      & & Ours     & 77.2 & 57.3 & 70.0 & 91.8 & 66.8 & 58.3 & 62.8 & 49.8 & 48.6 & 54.8 & 50.3 & 72.1 \\
      \cmidrule(l){3-15}
      & \multirow{2}{*}{Gemini 3 Flash~\citep{gemini3_flash}}
        & Baseline & 59.7 & 52.9 & 65.7 & 64.9 & 49.1 &  10.2 & 30.8 & 38.5 & 39.3 & 28.6 & 36.2 & 51.9 \\
      & & Ours     & 80.7 & 63.6 & 72.1 & 85.4 & 54.0 & 34.5 & 44.8 & 69.3 & 51.4 & 47.6 & 50.3 & 59.8 \\
      \cmidrule(l){3-15}
      & \multirow{2}{*}{Gemini 3.1 Pro~\citep{gemini31_pro}}
        & Baseline & 72.0 & 58.9 & 70.3 & 70.1 & 54.3 & 19.2 & 37.8 & 42.4 & 41.1 & 47.6 & 43.0 & 52.1 \\
      & & Ours     & 80.8 & \textbf{72.0} & \textbf{77.0} & 91.6 & 57.4 & \textbf{62.6} & 59.8 & \textbf{85.9} & 49.5 & 50.0 & 49.7 & 70.9 \\
      \cmidrule(l){3-15}
      & \multirow{2}{*}{Claude Sonnet 4.5~\citep{claude_sonnet_45}}
        & Baseline & 67.9 & 45.9 & 58.7 & 88.3 & 29.1 & 17.4 & 23.6 & 17.1 & 22.4 & 16.7 & 20.8 & 48.1 \\
      & & Ours     & 75.3 & 60.1 & 72.8 & 91.2 & 39.6 & 31.9 & 35.4 & 42.4 & 42.1 & 40.5 & 41.6 & 71.5 \\
      \cmidrule(l){3-15}
      & \multirow{2}{*}{Claude Sonnet 4.6~\citep{claude_sonnet_46}}
        & Baseline & 69.3 & 51.0 & 60.3 & 86.0 & 40.4 & 24.7 & 33.0 & 40.5 & 24.3 & 21.4 & 23.5 & 49.3 \\
      & & Ours     & \textbf{86.0} & 62.4 & 64.7 & 89.0 & 66.0 & 53.6 & 58.6 & 48.3 & 42.1 & 50.0 & 44.3 & 74.3 \\
      \cmidrule(l){3-15}
      & \multirow{2}{*}{Claude Opus 4.5~\citep{claude_opus_45}}
        & Baseline & 72.5 & 49.4 & 63.8 & 87.3 & 48.3 & 24.3 & 37.0 & 41.0 & 30.8 & 31.0 & 30.9 & 49.8 \\
      & & Ours     & 78.9 & 62.0 & 69.8 & 90.3 & 59.2 & 41.3 & 50.8 & 50.2 & 49.5 & 50.0 & 49.7 & 73.8 \\
      \cmidrule(l){3-15}
      & \multirow{2}{*}{Claude Opus 4.6~\citep{claude_opus_46}}
        & Baseline & 76.2 & 55.8 & 64.4 & 89.9 & 62.3 & 30.2 & 47.2 & 42.0 & 36.4 & 42.9 & 38.3 & 51.2 \\
      & & Ours     & 80.2 & 66.0 & 71.8 & \textbf{92.2} & \textbf{72.8} & 57.9 & \textbf{65.8} & 53.7 & \textbf{55.1} & \textbf{59.5} & \textbf{56.4} & \textbf{82.3} \\
    \bottomrule
  \end{tabular}%
  }
  \par\vspace{3pt}
  \parbox{\textwidth}{\footnotesize\raggedright
    $^{\dagger}$SuperChem overall SOTA (38.5\%) predates recent text-only results; headline uses the text subset.}
  \vspace{-5pt}
\end{table}

\vspace{-10pt}
\section{Experiments}\label{sec:experiments}
\vspace{-5pt}

\subsection{Setup}\label{subsec:exp_setup}
\vspace{-5pt}

We evaluate 12 backbones, including 5 open-source models, i.e., Trinity-Large-Thinking~\citep{trinity2025}, Nemotron-3-Super-120B-A12B~\citep{nemotron3_2025}, INTELLECT-3.1~\citep{intellect31_2026}, GLM-4.5~\citep{glm45_2025}, and Qwen3-235B-A22B~\citep{qwen35_2026}, and 7 closed-source models (Claude Sonnet 4.5/4.6, Opus 4.5/4.6, GPT-5.4, Gemini 3 Flash, and Gemini 3.1 Pro). The evaluations are conducted on 8 representative benchmarks: SuperChem~\citep{superchem2025}, HLE-Gold (Bio+Chem)~\citep{hle2025}, HealthBench Hard~\citep{healthbench2025}, MedXpertQA~\citep{medxpertqa2024}, ProteinLMBench~\citep{proteinlmbench2024}, Medbullets~\citep{medbullets2024} (\texttt{op4} 4-option subset), BixBench~\citep{bixbench2025}, and LAB-Bench 2~\citep{labbench2_2025}.
For our mutual-evolve harness, we set the cohort size (number of solvers) equal to 4; the temperature schedule $\{\tau_1, \ldots, \tau_N\}$ is uniformly sampled from $[0.1, 0.9]$, the private-phase length $T=10$, the read interval $K=3$, the minimum tool-use depth $L_{\min}=10$, and the voting coefficient $\beta=0.1$.

\vspace{-10pt}
\subsection{Ablation Study}\label{subsec:ablation}
\vspace{-5pt}

We first provide an ablation study in Table~\ref{tab:ablation} to better understand the contributions of each design. Our toolkit includes six agent harnesses, including our proposed mutual-evolve harness, and six context-management strategies. As we can see, all agent harnesses can significantly boost the performance of the base model, indicating the importance of building harnesses for deep research. Among the six harnesses, our mutual-evolve harness achieves the best performance. Further, in terms of context management, we evaluate all implemented strategies on Sonnet 4.5 and find that planning (which records intermediate state worth carrying forward, such as key facts and unresolved subgoals) paired with \emph{summarization} (which compresses only newly aged trace segments once the running input crosses a length threshold) is the most robust combination across HLE-Gold and BixBench. We therefore fix this pair as the default context-management configuration in all subsequent experiments. For our mutual-evolve harness, we perform a sensitivity analysis of the hyperparameters: number of solvers $N$, private phase, and shared phase. The results confirm our hyperparameter selection. Furthermore, we can clearly see that fewer private phases achieve lower performance, possibly because findings provided in the early stages of agent research may be of low quality and thus mislead other agents' research, indicating the effectiveness of introducing the private phase. Both the final confirmation and weighted voting can boost performance by further exploiting the information shared and written in the global workspace when agents are performing deep research.

\vspace{-5pt}
\subsection{Comparison with State-of-the-Art}
\vspace{-5pt}

Based on the ablation study in Section~\ref{subsec:ablation} and considering the balance between accuracy and reasoning cost, we adopt \textsc{Light Mutual-Evolve} as the default harness in BioMedArena and use it to boost a range of backbones for comparison against the published state-of-the-art on eight biomedical benchmarks. Table~\ref{tab:main_results} reports the detailed results.
We can see that BioMedArena lifts these backbones to exceed the SOTA on the eight benchmarks.
For example, on HLE-Gold, BioMedArena boosts Claude Opus 4.6 to reach {56.4\%} accuracy, outperforming the SOTA of 46.8\% by +9.6 pp; Gemini 3 Flash + Ours (50.3\%) and Gemini 3.1 Pro + Ours (49.7\%) also clear this SOTA. On {BixBench}, Gemini 3.1 Pro + Ours reaches {85.9\%}, beating the GPT-5.5 SOTA of 80.5\% by +5.4\,pp. On {SuperChem}, Claude Opus 4.6 + Ours reaches {72.8\%} on the text-only subset, a +9.6\,pp lift over the MiroThinker-1.7\,\&\,H1 SOTA of 63.2\%; counting the harder image questions it reaches 65.8\% overall, against an older overall SOTA of 38.5\%~\citep{superchem2025} that predates recent text-only results and is not directly comparable. Table~\ref{tab:main_results} reports the text/image breakdown, with the image subset showing the largest harness effect (e.g., Gemini 3.1 Pro rises from 19.2\% to 62.6\%, a +43.4\,pp lift). On {LAB-Bench\,2} (the 7 text-only subsets), Claude Opus 4.6 + Ours reaches {82.3\%}, beating the SOTA of 80.0\% by +2.3\,pp.

\begin{figure}[t]
  \centering
  \includegraphics[width=0.85\textwidth]{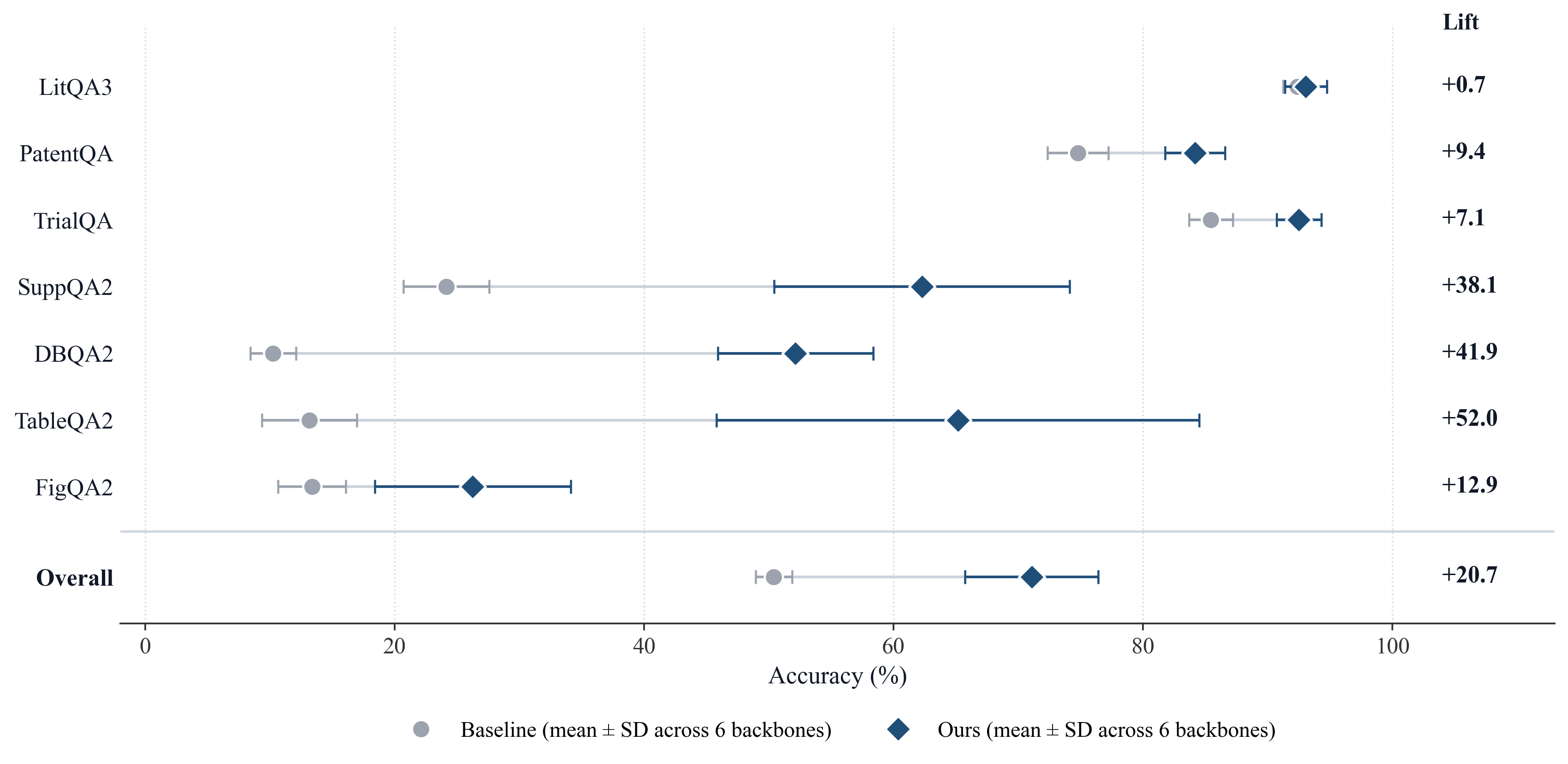}
   \vspace{-5pt}
 \caption{Per-subset accuracy on the LAB-Bench\,2 7-subset text-only subset, averaged across 6 backbones (2 Gemini, 4 Claude). Markers show mean accuracy under Baseline (gray circles) and Ours (dark diamonds); error bars indicate $\pm 1$ SD across backbones; the rightmost column reports the per-subset lift (Ours $-$ Baseline) in percentage points.}
 \label{fig:labbench2_plot}
 \vspace{-5pt}
\end{figure}

\vspace{-5pt}
\subsection{LAB-Bench\,2 per-subset breakdown}\label{subsec:per_bench_breakdown}
\vspace{-5pt}

Figure~\ref{fig:labbench2_plot} unpacks the LAB-Bench\,2 Overall number into its 7 text-only subsets, plotting per-subset accuracy averaged across 6 backbones (2 Gemini, 4 Claude) under Baseline and Ours, with error bars showing $\pm 1$ SD across backbones and the lift annotated on the right. The literature-style subsets at the top (LitQA3, PatentQA, TrialQA) already sit between 75\% and 93\% under Baseline, so the harness yields only modest lifts of $+0.7$ to $+9.4$\,pp. The retrieval-grounded subsets in the middle (SuppQA2, DBQA2, TableQA2) drive the bulk of the improvement: long horizontal gaps separate the gray Baseline markers (10--25\%) from the dark Ours markers (50--65\%), with lifts of $+38.1$, $+41.9$, and $+52.0$\,pp---these subsets require pulling and parsing structured external evidence that Baseline cannot reach. FigQA2 at the bottom remains the hardest subset ($\sim$26\% under Ours, $+12.9$\,pp lift), reflecting the limit of our text-only tool surface on figure-grounded questions. Aggregated, the Overall row rises from $\sim$50\% to $\sim$71\% ($+20.7$\,pp), driven primarily by the three retrieval-grounded subsets rather than uniform improvement.

\section{Conclusion}\label{sec:conclusion}

We presented \textsc{BioMedArena}, an open-source toolkit that turns biomedical deep-research agent evaluation from repeated, system-specific engineering into shared infrastructure, registering 166 benchmarks, 75 typed tools, 6 agent harnesses, and 6 context-management strategies behind a unified provider abstraction reachable from a few-line model adapter. We further proposed \textsc{Mutual-Evolve}, a harness in which parallel solvers share intermediate findings through a typed Global Workspace and aggregate answers via contribution-weighted voting. A 12-backbone matrix study under one fixed harness establishes new state-of-the-art results on all 8 benchmarks, with an average lift of +15.01\,pp. The released toolkit provides a shared, reproducible evaluation environment under which future biomedical deep research agents can be compared head-to-head from day one.

\bibliographystyle{plainnat}
\bibliography{references}

\newpage
\appendix

\newpage

\section{Limitations}
First, our tool surface is largely text-based, and benchmarks whose answers depend on grounded images or large-table retrieval (e.g. the FigQA2 and TableQA2 subsets of LAB-Bench~2) retain visible headroom in our results. 
Second, the harness, tool registry, and scoring policy are fixed across rows, but the LLM-as-judge component still introduces some variance in open-ended tasks, and Mutual-Evolve's hyperparameters are tuned on a held-out subset rather than optimised per benchmark.

\section{Broader Impact}
By turning per-paper evaluation infrastructure into a shared, open-source toolkit, BioMedArena lowers the engineering barrier to biomedical agent research and allows smaller laboratories to run rigorous head-to-head comparisons with a few-line provider adapter. Concentrated attention on a fixed benchmark set may encourage over-optimisation, which we mitigate by registering 166 benchmarks so that the community can rotate the evaluation surface as benchmarks saturate.

\section{Ethics Considerations}
The involved benchmark datasets are all open-source. We only use public data secondarily and do not recruit any human research participants or create new data for this study.
BioMedArena is an evaluation toolkit, not a clinical system, and the results should not be interpreted as endorsements for clinical decision-making. No underlying patient-level data are redistributed; users access each benchmark under its original licence, and tool calls to external biomedical APIs are logged for auditability. 

\section{Comparison with Existing Evaluation Systems}\label{app:comparison}

\begin{table}[h]
\caption{Architectural comparison across eight framework-level capabilities that determine whether deep research agent results are reproducible and comparable across models: each capability removes one source of cross-paper variance (harness, tool surface, scoring, or backbone interface). \checkmark{} indicates out-of-the-box support; ($\times$) indicates absent or only ad-hoc. \emph{Context manager}: trace compression beyond raw concatenation. \emph{Multi-harness}: multiple deep-research harnesses selectable. \emph{Trace logging}: tool calls and responses persisted at framework level. \emph{Tool routing}: per-benchmark or per-query tool filtering. \emph{Custom tool}: user-extensible typed registry. \emph{Multi-domain}: clinical, omics, chemistry, and literature coverage. \emph{Multi-scoring}: MCQ, open-ended judge, code execution, and structured match. \emph{Multi-backbone}: unified abstraction over open- and closed-weight LLMs.}
  \label{tab:related}
  \centering
  \footnotesize
  \setlength{\tabcolsep}{5pt}
  \renewcommand{\arraystretch}{1.15}
  \begin{tabular}{lcccccccc}
    \toprule
    System
      & \shortstack{Context\\manager}
      & \shortstack{Multi-\\harness}
      & \shortstack{Trace\\logging}
      & \shortstack{Tool \\routing}
      & \shortstack{Custom\\tool}
      & \shortstack{Multi-\\domain}
      & \shortstack{Multi-\\scoring}
      & \shortstack{Multi-\\backbone} \\
    \midrule
    Biomni~\citep{biomni2025}                    & \(\times\) & \(\times\) & \(\times\) & \(\times\) & \checkmark & \checkmark & \(\times\) & \(\times\) \\
    MedAgentBench~\citep{medagentbench2024}      & \(\times\) & \(\times\) & \(\times\) & \(\times\) & \(\times\) & \(\times\) & \(\times\) & \(\times\) \\
    AgentClinic~\citep{agentclinic2024}          & \(\times\) & \(\times\) & \(\times\) & \(\times\) & \(\times\) & \(\times\) & \(\times\) & \(\times\) \\
    MedHELM~\citep{medhelm2025}                  & \(\times\) & \(\times\) & \(\times\) & \(\times\) & \(\times\) & \checkmark & \checkmark & \checkmark \\
    HELM~\citep{helm2023}                        & \(\times\) & \(\times\) & \(\times\) & \(\times\) & \(\times\) & \(\times\) & \checkmark & \checkmark \\
    \texttt{lm-eval-harness}~\citep{lmevalharness} & \(\times\) & \(\times\) & \(\times\) & \(\times\) & \(\times\) & \(\times\) & \checkmark & \checkmark \\
    \midrule
    \rowcolor{gray!12}
    \textbf{BioMedArena} (ours)                  & \checkmark & \checkmark & \checkmark & \checkmark & \checkmark & \checkmark & \checkmark & \checkmark \\
    \bottomrule
  \end{tabular}
\end{table}

\section{Benchmark Details}\label{app:benchmark_details}

\begin{table}[htbp]
\centering
\caption{The 8 biomedical benchmarks used in our headline experiments, with $N$ the number of evaluated questions.}
\label{tab:benchmarks}
\scriptsize
\setlength{\tabcolsep}{4pt}
\renewcommand{\arraystretch}{1.15}
\begin{tabular}{@{}lrllll@{}}
\toprule
\textbf{Benchmark} & \textbf{$N$} & \textbf{Domain} & \textbf{Capability} & \textbf{Answer Format} & \textbf{Scoring} \\
\midrule
\textbf{MedXpertQA Text}              & 2{,}450      & Clinical medicine  & Expert MCQ reasoning       & 10-opt MCQ       & Rule         \\
\midrule
\textbf{Medbullets op4}               & 308          & Clinical medicine  & USMLE-style reasoning      & MCQ              & Rule         \\
\midrule
\textbf{ProteinLM Bench}              & 944          & Proteins           & Domain knowledge           & MCQ              & Rule         \\
\midrule
\textbf{HealthBench Hard}             & 1{,}000      & Health dialog      & Open-ended generation      & Open-text        & Rubric judge \\
\midrule
\textbf{BixBench}                     & 205          & Comp.\ biology     & Comp.\ biology reasoning   & MCQ (adapted)    & Rule         \\
\midrule
\textbf{HLE-Gold Bio/Chem (verified)} & \textbf{149} & Bio/Chem/Med       & Expert frontier reasoning  & MCQ + open       & Rule + judge \\
\quad Bio                             & 107          & Biology/Medicine   &                            &                  &              \\
\quad Chem                            & 42           & Chemistry          &                            &                  &              \\
\midrule
\textbf{SuperChem}                    & \textbf{500} & Chemistry          & Multimodal chem.           & MCQ              & Rule         \\
\quad Text                            & 265          & Chemistry          & Text chem.\ reasoning      & MCQ              &              \\
\quad Image                           & 235          & Chemistry          & Multimodal chem.\ reasoning & MCQ + image     &              \\
\midrule
\textbf{LAB-Bench 2}                  & \textbf{821} & Lab research       & Retrieval / grounding      & Open-text        & Regex / validator \\
\quad LitQA3                          & 168          & Literature         & Literature comprehension   &                  &              \\
\quad PatentQA                        & 121          & Patents            & Patent search              &                  &              \\
\quad TrialQA                         & 120          & Clinical trials    & Trial lookup               &                  &              \\
\quad DBQA2                           & 86           & Databases          & Structured retrieval       &                  &              \\
\quad SuppQA2                         & 125          & Supp.\ material    & Document grounding         &                  &              \\
\quad FigQA2                          & 101          & Figures            & Figure QA (text-only)      &                  &              \\
\quad TableQA2                        & 100          & Tables             & Table QA (text-only)       &                  &              \\
\bottomrule
\end{tabular}
\end{table}

\section{Context Management}

\begin{table}[htbp]
\centering
\caption{The six context-management strategies in the BioMedArena, organized by methodological function. The Core idea column states what the strategy maintains, compresses, removes, or recovers; the What it protects against column states the failure mode it addresses; the Setting column reports the default in our headline harness and whether the strategy is ablatable. Each strategy emits structured trace records under a unified schema, supporting decomposable post-hoc ablation from released artifacts.}
\label{tab:context-strategies}
\footnotesize
\setlength{\tabcolsep}{4pt}
\renewcommand{\arraystretch}{1.3}
\begin{tabular}{@{}lp{4cm}p{4cm}p{3cm}@{}}
\toprule
\textbf{Strategy} & \textbf{Core idea} & \textbf{What it protects against} & \textbf{Setting} \\
\midrule
\textbf{Planning}         & Maintain compact working notes of intermediate findings, unresolved subgoals, and accumulated evidence                          & Loss of useful intermediate state across multi-step tool use                                            & Default in \texttt{light} harness; ablatable \\
\addlinespace
\textbf{Memory}           & Persist salient facts, entities, and prior findings to an external store and re-inject relevant entries on demand                                            &   Loss of cross-turn or cross-session knowledge that no longer fits in the live context                 &  Default in \texttt{light} harness; ablatable\\
\addlinespace
\textbf{Summarization}      & Compress earlier dialogue, tool outputs, and evidence into shorter summaries while preserving recent context                    & Long traces crowding out useful context or exceeding provider limits                                    & Optional; length-triggered \\
\addlinespace
\textbf{Clearing}           & Replace stale or verbose payloads, such as old tool outputs, reasoning traces, or media, with compact placeholders               & Low-value context bulk from long tool results, repeated outputs, or large media payloads                & Optional; horizon-triggered \\
\addlinespace
\textbf{Truncation}         & Keep task-relevant context through sliding windows, first-last retention, or token budgets                                       & Context overflow when summarization or clearing is insufficient                                         & Optional; budget-triggered \\
\addlinespace
\textbf{Rollback}           & Remove the most recent low-value assistant/tool turn and insert corrective guidance                                              & Repeated search queries, repeated URL scrapes, tool errors, and early loop formation                    & Optional; loop-triggered \\

\bottomrule
\end{tabular}
\end{table}

\begin{figure}[t]
  \centering
  \includegraphics[width=0.8\textwidth]{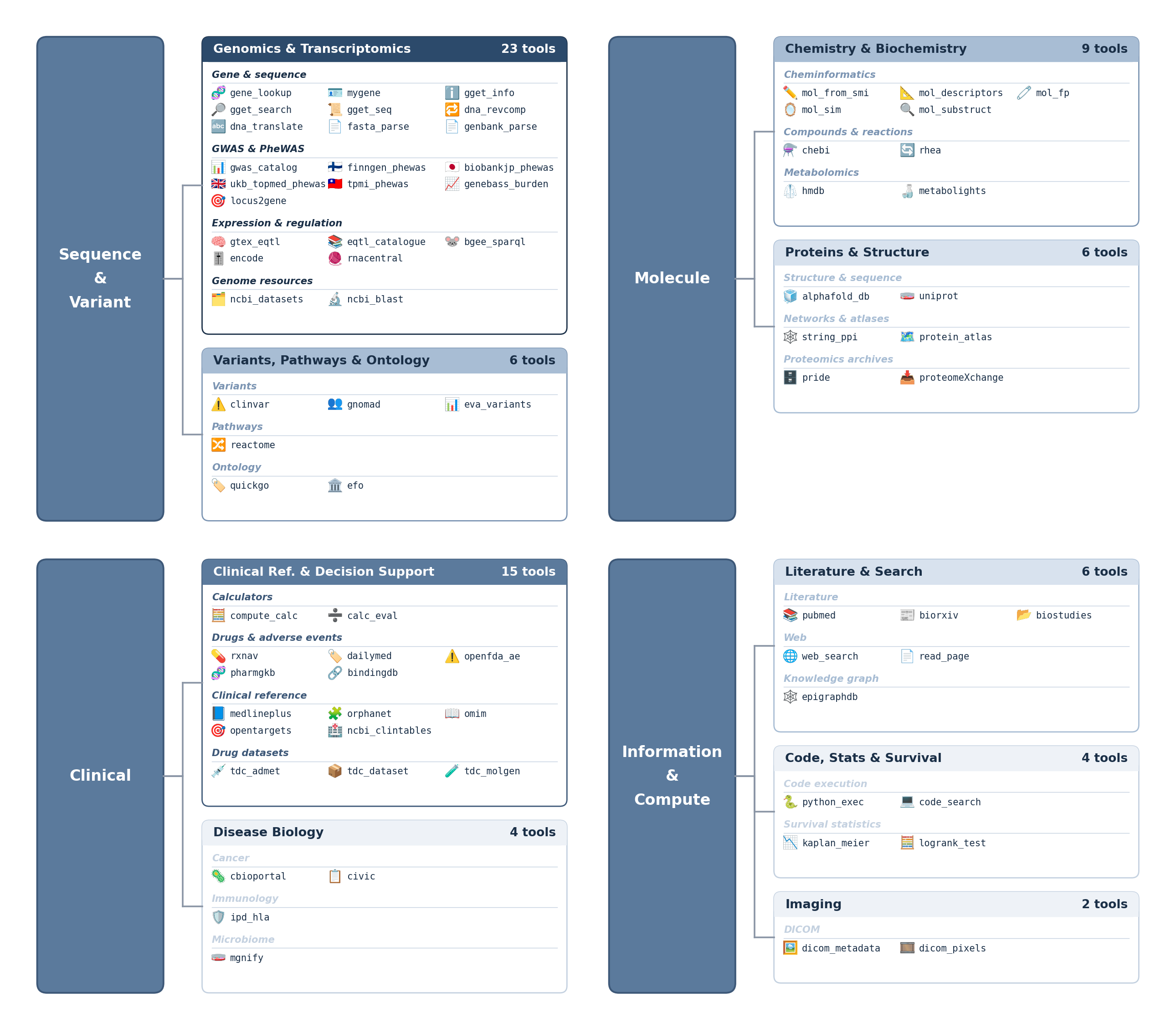}
  \vspace{-10pt}
  \caption{Tool registry organized by biomedical skill family.
  The 33 category tags group into 9 families with 75
  tools. Counts are
  non-exclusive because a tool may carry multiple category tags.}
  \label{fig:tool_overview}
\end{figure}

\section{Mutual-Evolve Harness Algorithm }
\label{sec:me_algorithm}

The framework-level orchestration is given as Algorithm~\ref{alg:mutual_evolve}, and the parallel agent loop as Algorithm~\ref{alg:solver_rollout}.

\begin{algorithm}[t]
\scriptsize
\caption{\textsc{Mutual-Evolve}}
\label{alg:mutual_evolve}
\begin{algorithmic}[1]
\Require question $q$; tool set $\mathcal{T}$; $N$ solvers with temperatures $\{\tau_i\}$; private-phase length $T$, read interval $K$, minimum tool-use depth $L_{\min}$; voting coefficient $\beta$
\Ensure prediction $\hat{a}$
\State $\mathcal{W} \gets$ new GlobalWorkspace; $\mathcal{B} \gets$ new Barrier$(N)$
\For{$i = 1$ \textbf{to} $N$ \textbf{in parallel}}
    \State $(a_i, \mathrm{resp}_i, \mathrm{status}_i) \gets \textsc{SolverRollout}(q, \mathcal{T}, \tau_i, \mathcal{W}, \mathcal{B}, T, K, L_{\min}, i)$
\EndFor
\State $\mathcal{S} \gets \{i : \mathrm{status}_i = \texttt{completed}\}$
\For{$i \in \mathcal{S}$ \textbf{in parallel}} \Comment{Final Confirmation}
    \State $\mathrm{resp}_i \gets \textsc{LLM.Chat}(\mathrm{ctx}_i \oplus \mathcal{W}.\textsc{ReadAll}(), \texttt{tools} = \emptyset)$
    \State $a_i \gets \textsc{ExtractAnswer}(\mathrm{resp}_i)$
\EndFor
\For{$i \in \mathcal{S}$}
    \State $w_i \gets 1 + \beta \cdot \mathcal{W}.\textsc{Count}(i)$ \Comment{$\mathcal{W}.\textsc{Count}(i)$: \# entries written by solver $i$}
\EndFor
\State $\hat{a} \gets \arg\max_a \sum_{i \in \mathcal{S},\, a_i = a} w_i$ \Comment{ties: first-appearing}
\State \Return $\hat{a}$
\end{algorithmic}
\end{algorithm}

\section{Experiments Compute Resources}\label{app:compute}

All open-source backbones (Trinity-Large-Thinking, Nemotron-3-Super-120B-A12B, INTELLECT-3.1, GLM-4.5, and Qwen3-235B-A22B) are served locally via vLLM on a single node with 8$\times$NVIDIA H200 GPUs (141\,GB HBM3e each), using tensor parallelism across the 8 GPUs. Closed-source backbones (Claude Sonnet 4.5/4.6, Claude Opus 4.5/4.6, GPT-5.4, Gemini 3 Flash, and Gemini 3.1 Pro) are accessed through their official provider APIs. The LLM judge used in the scoring router is also accessed via API. Per-call latency, token usage, and cost are recorded for every model invocation in the released run logs.

\end{document}